# Analog and Multi-modal Manufacturing Datasets Acquired on the Future Factories Platform

Ramy Harik, Fadi El Kalach, Jad Samaha, Devon Clark, Drew Sander, Philip Samaha, Liam Burns, Ibrahim Yousif, Victor Gadow, Theodros Tarekegne, Nitol Saha

Department of Mechanical Engineering, University of South Carolina
Columbia, South Carolina, USA, 29201

Corresponding Author: harik@mailbox.sc.edu

## Abstract

Two industry-grade datasets are presented in this paper that were collected at the Future Factories Lab at the University of South Carolina on December 11th and 12th of 2023. These datasets are generated by a manufacturing assembly line that utilizes industrial standards with respect to actuators, control mechanisms, and transducers. The two datasets were both generated simultaneously by operating the assembly line for 30 consecutive hours (with minor filtering) and collecting data from sensors equipped throughout the system. During operation, defects were also introduced into the assembly operation by manually removing parts needed for the final assembly. The datasets generated include a **time series analog dataset** and the other is a **time series multi-modal dataset** which includes images of the system alongside the analog data. These datasets were generated with the objective of providing tools to further the research towards enhancing intelligence in manufacturing. Real manufacturing datasets can be scarce let alone datasets with anomalies or defects. As such these datasets hope to address this gap and provide researchers with a foundation to build and train Artificial Intelligence models applicable for the manufacturing industry. Finally, these datasets are the first iteration of published data from the future Factories lab and can be further adjusted to fit more researchers' needs moving forward.

## I. Problem

During the past century, the United States transitioned from being the foremost autonomous and interconnected manufacturer to become heavily reliant on other nations. This shift presents genuine risks and obstacles for the manufacturing industry. The McKinsey Global Institute (MGI) conducted research highlighting key components, rooted in Industry 4.0, that are necessary to improve manufacturing and increase efficiency. These include optimized processes, improved asset utilization, better supply chain management, and more efficient inventory practices, among other factors [1]. In the era of Industry 4.0, a transformative wave of cutting-edge technologies and innovations is reshaping the manufacturing landscape. This marked the beginning of data-driven manufacturing [2], a pivotal shift centered on fostering continuous data exchange and harnessing its abundance across applications that hold the potential to significantly enhance productivity and efficiency within the sector. The integration of advanced sensors, proficient in capturing multifaceted aspects of processes, has played a pivotal role in facilitating the widespread generation of data. The rise of Artificial Intelligence (AI) concepts has notably impacted critical domains such as predictive maintenance, quality control [3], worker safety [4], and process optimization, which underscores the heightened importance of meaningful data. As a result, there is an escalating necessity for comprehensive industrial datasets—versatile repositories that can serve as the training ground for a diverse array of task-specific AI algorithms.

Yet navigating the landscape of industrial dataset generation is a multifaceted challenge marked by several intricate hurdles. Foremost among these challenges is the concern of data privacy and security. Companies, protective of their proprietary processes, often hesitate to share datasets to safeguard confidentiality within their manufacturing facilities. Adding to the complexity is the inherent intricacy of industrial settings. Manufacturing processes, with their numerous variables, pose difficulties in comprehensive representation, especially when sensors capable of capturing every aspect of the process are not available. The variability of datasets, particularly those incorporating anomalies, proves to be another formidable obstacle. Introducing anomalies during manufacturing operations is often impractical due to process disruptions and associated costs, rendering the generation of datasets with maximum variability and authentic representation of anomalies challenging.

Moreover, the sheer scale of industrial datasets, exacerbated by the computational demands of accommodating large volumes of data, adds another layer of complexity. Manufacturing processes, generating copious amounts of data, necessitate substantial computational resources. The challenge lies in efficiently gathering multi-modal data from various sources throughout entire manufacturing processes, a task that can be troublesome and, in some cases, unfeasible without the proper resources. Collectively, these challenges underscore the intricate nature of industrial dataset generation, emphasizing the scarcity of such datasets within the public domain.

To tackle these issues, the Future Factories team at the University of South Carolina has assembled a manufacturing testbed that utilizes the same industrial standards used in manufacturing facilities. This testbed is used for manufacturing research across multidisciplinary topics. As such this paper will introduce all the components of this testbed and outline the manufacturing process implemented within the testbed. This paper will also introduce two datasets, an Analog and Multi-modal dataset which are described thoroughly in the remainder of this paper. The multi-modal dataset represents a novel dataset which synchronizes live image frames of the assembly line with the analog data generated during the same instance.

# II. Experimental Set-up

To clearly define the dataset, the experimental set up must be outlined to illustrate the insights behind the generated data. This section will be split up into three subsections. The first subsection lists all the equipment (actuators and transducers) that are used in the testbed, the second subsection will describe the naming convention of the dataset features. Finally, the last subsection will describe the assembly process that takes place within the testbed.

## a) Equipment

The layout of the Future Factories Testbed is shown in Figures 1 and 2. The testbed contains five Yaskawa industrial arms, a conveyor system, and a material handling station.

**Robotic Arms**

The testbed has five Yaskawa six-axis robots. Two of which are Yaskawa HC10s and three GP8s which are controlled by YRC1000 and YRC1000micro robot controllers. Due to their high speed and precision in repetitive tasks, these robots can collaboratively assemble different products. The two HC10s are utilized to streamline material intake and output processes while the GP8s are used to assemble and disassemble products at their separate stations. Each robot has a custom designed, 3D printed gripper attached as its end effector.

**Conveyor System**

The FF testbed features a four-conveyor system interlinked to form a complete loop around all the robots. The C4N Conveyor Belts and Stands play a crucial role in transporting the products across the different stations within the testbed. These conveyors are controlled by the

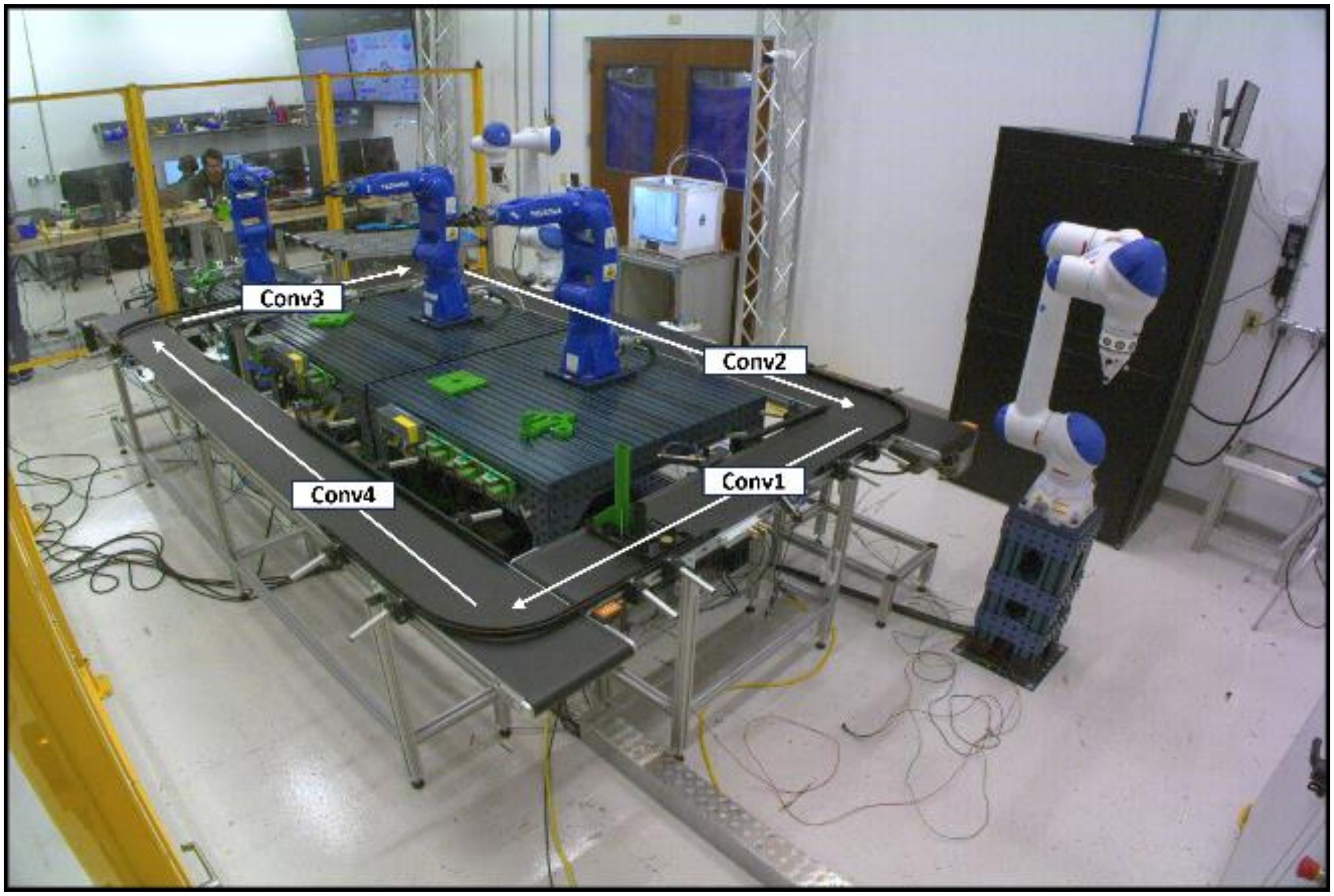

*Figure 1: Future Factories Testbed Set Up (View 1)*

Sinamics GS120 Variable Frequency Drives (VFDs), which are connected to the Programmable Logic Controller. The conveyor system is essential for facilitating collaboration among the different robotic arms.

## Programmable Logic Controller (PLC)

The communication between devices and machines in the testbed is made possible through the Siemens S7-1500 PLC. The PLC code is written using Siemens Totally Integrated Automation (TIA) Portal engineering software, where the commissioning and management of device layout occurs. TIA is also used to write the control logic deployed onto the PLC required to carry out the assembly process. The PLC communicates with connected devices using various communication protocols. Three Distributed I/O devices are connected to the PLC, serving as extended I/O modules which can be expanded modularly as needed. The PLC also employs the Profinet communication to communicate with the I/O modules, the robot controllers, and the Conveyor VFDs.

## Sensors

The two types of sensors utilized in this dataset are mounted onto the end effectors of the robots. These sensors are utilized to derive insights on the status of the assembly process as the generated signal differs when different material handling errors occur.

### *Potentiometer*

The primary function of the potentiometer is to relay information regarding the status of the gripper. The Sensata-BEI potentiometer produces distinct values when the gripper is closed compared to when it is open. Consequently, the generated value serves as an indicator of whether the gripper is in an open or closed state. The resistive sensor generates a voltage, which is directly input into our PLC I/O modules.

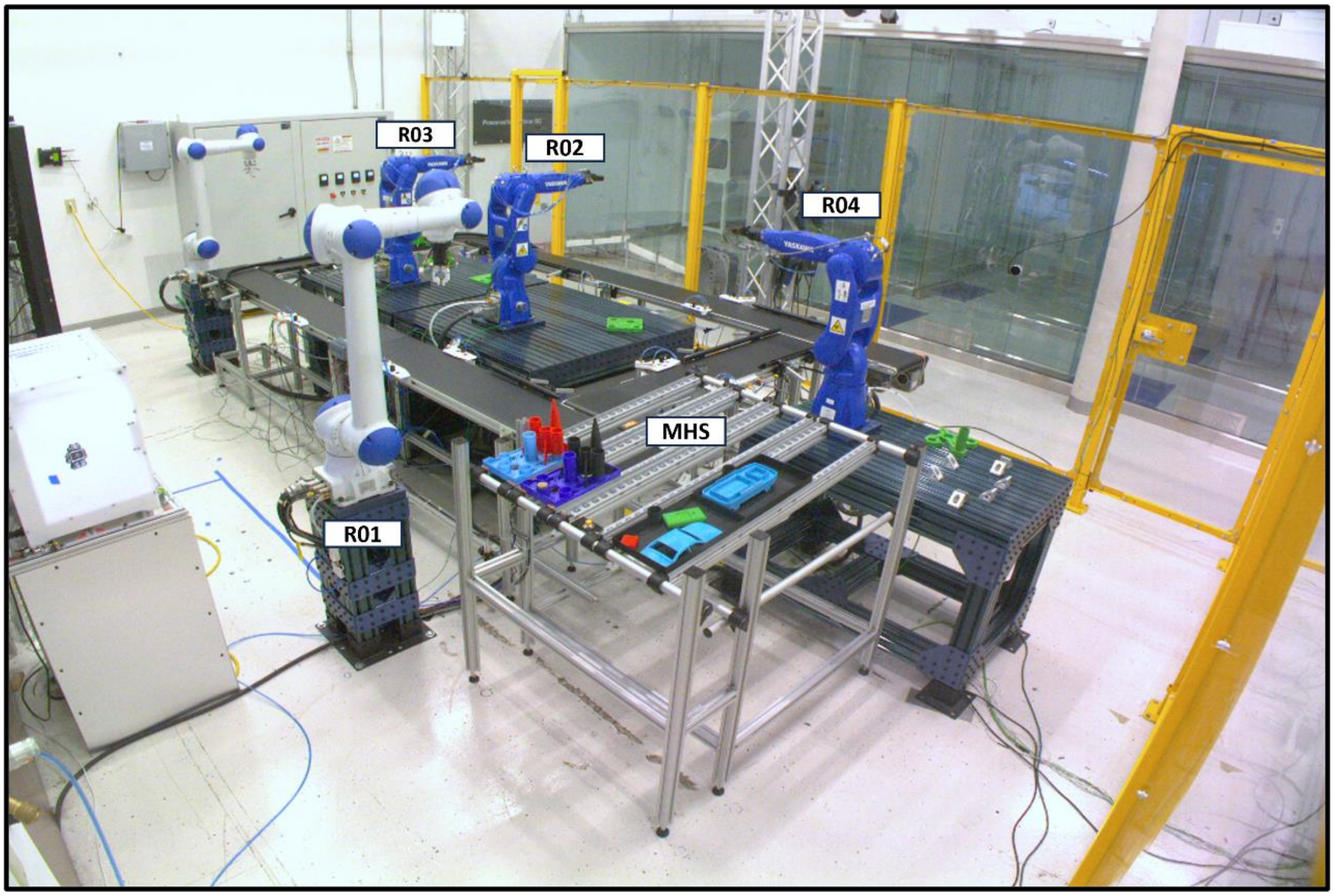

*Figure 2: Future Factories Testbed Set Up (View 2)*

*Load Cell*

The TE Connectivity amplified load cell gauges the force or pressure applied to it. In the testbed, its main role is to determine whether the gripper is presently holding an object. When the robot's gripper is closed and holding an object, the generated load cell value increases proportionally with the increased pressure. Whereas, when there is no object, the value remains minimal. The load cell is directly connected to the PLC I/O module.

## b) Naming Convention

In order to standardize the naming convention, each asset was given a unique identifier based on the chronological order in which that asset is used. As such, robots are noted as R01 (i.e. Robot 01), R02, R03, and R04. The stations positioned on the conveyors are denoted as Stop1, Stop2, Stop3, Stop4, and Stop5. Each identifier can be seen in Figures 1 and 2.

## c) Assembly Process

The testbed was created to carry out an assembly operation utilizing robots and conveyors. The process starts with R01 picking up a tray with the unassembled rocket from the material handling station. The robot places the tray on the conveyor which transports the tray onto the first station at R02. At that point, R02 will pick up the two body pieces from the tray and place them onto its separate assembly station. After that is done, the conveyor transports the tray down the line till it reaches the second station at R03. At this station R03 places the base of the rocket in its separate assembly station. R02 will then hand off the two body pieces to R03 to be assembled on top of the base. Finally, R03 picks up the nose cone from the tray, places it on the now fully assembled rocket before returning the rocket to the tray. Once the rocket is fully assembled, the conveyor system transports the tray to the other side of the manufacturing system till it reaches the third station. At that point R04 picks up the tray and places it on its separate station. R04 then disassembles the rocket piece by piece and returns the disassembled rocket onto the tray and places it onto the material handling station for R01 to introduce it back into the system and the whole process repeats itself. This full process is what is referred to as a cycle in this dataset.

# III. Data Metrics

## a) Analog Dataset

The dataset features 30 hours of operation of the assembly and disassembly process. Upon completion of the experimental run, the different sensor values shown in the Appendix were downloaded and sorted into multiple CSV files. These CSV files were sorted based on the asset they pertain to (i.e. R01's signals in R01_Data.csv) as seen in Figure 4. On top of that, the data was also cleaned. Throughout the 30 hours of operation, the testbed experienced a bit of downtime leading to meaningless data. The time range during which the testbed was not operating were filtered out of the dataset such that the final dataset consists only of 325 complete cycles.

Throughout the 30 hours, some anomalies were also simulated by having team members manually remove rocket pieces from the tray so as to manufacture defective products. These anomalies are classified into three categories which signify how many pieces are missing from the four-piece rocket:

- NoNoseCone
- NoBody2,NoNose
- NoBody1,NoBody2,NoNose

These anomalies are annotated in the analog dataset where annotations are associated with the cycles (i.e. Cycle 1 is not anomalous or Cycle 50 has the anomaly NoNoseCone). Apart from the annotation and lack of image data, the other pivotal difference between this dataset and the multi-modal dataset is the acquisition rate of 10 Hz.

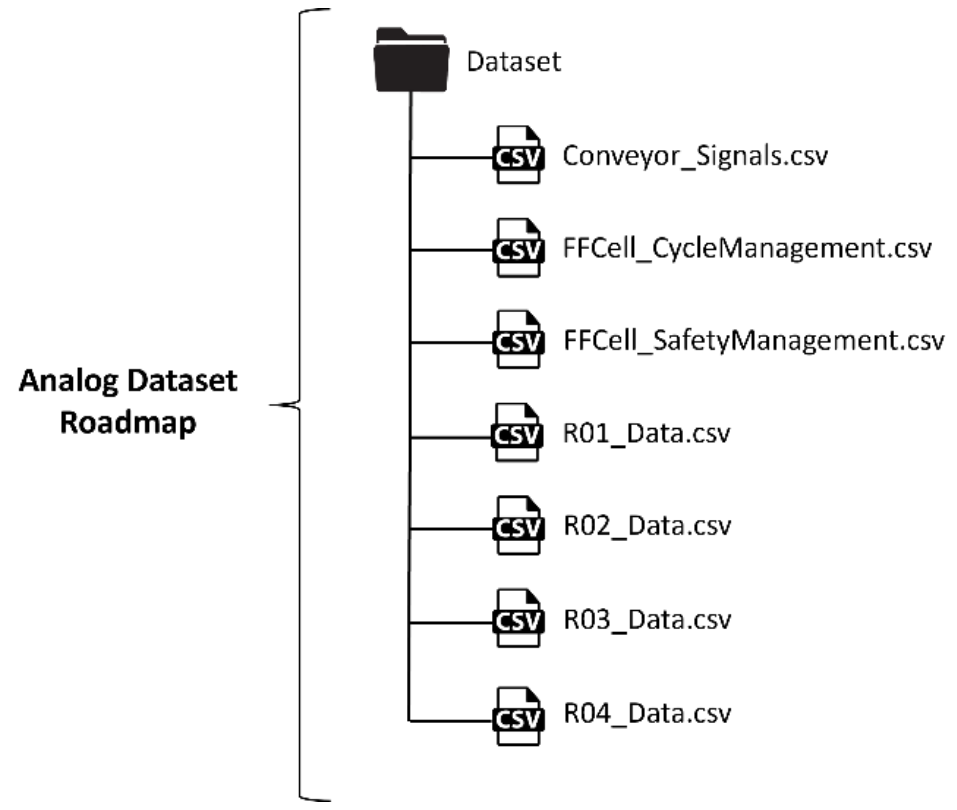

*Figure 2: Configuration of the Analog Dataset*

## b) Multi-modal dataset

Similar to the Analog dataset, the multi-modal dataset was generated from the assembly and disassembly of a rocket prototype for identical runtime and circumstances. The sensor values involved in this dataset are shown in the Appendix. Additionally, this dataset includes synchronized image data acquired from two cameras mounted on either side of the testbed, continuously capturing scenes of the operations. Consequently, the acquisition rate dropped, spanning from 2 to 3 data samples per second and resulting in a total of 166,000 records throughout the entire runtime.

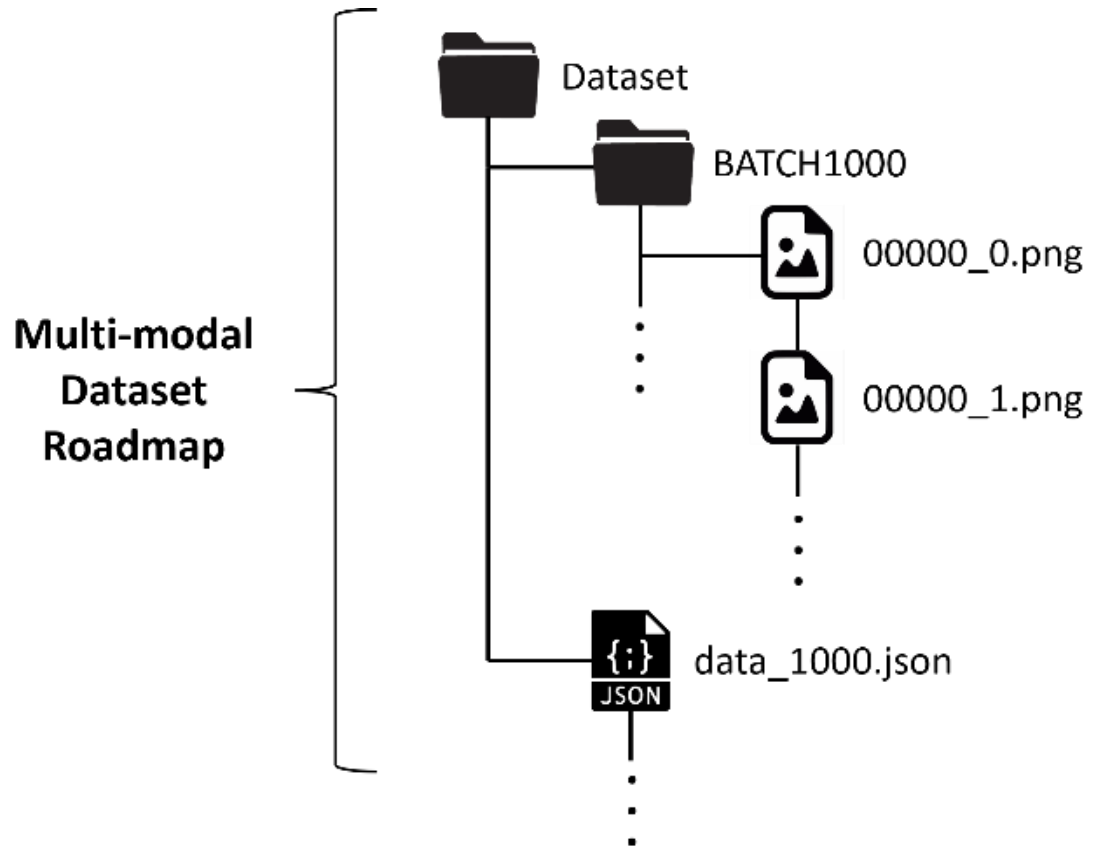

*Figure 3: Configuration of the Multi-modal Dataset*

While the analog dataset is tabularly structured and stored in CSV files, the multi-modal dataset entails a different structure. As shown in Figure 3, the images are divided into batches of 1000 samples of both camera viewpoints and stored in separate folders. For each batch, an equivalent JSON file can be found containing the synchronized sensor values and their corresponding image paths. Given the number of records, the dataset folder consists of a total of 166 images batch folders and their respective JSON files.

# IV. Concluding Remarks

This dataset is made widely available for researchers to use for their individual research. At this time, this is the first dataset published by the neXt Future Factories team at the University of South Carolina, and the data is organized in a fashion that meets the requirements of our research team. Researchers are welcome to reach out for adjustments or more data for the next iterations through emailing the Principal Investigator Professor Ramy Harik at harik@mailbox.sc.edu.

# Download Links

The **analog dataset** is downloadable using the following link:

- Analog Link 1 [https://www.kaggle.com/datasets/ramyharik/ff-2023-12-12-analog-dataset]

The **multi-modal dataset** is divided into 6 different downloadable links:

- Multi-Modal Link 1 [www.kaggle.com/datasets/ramyharik/ff-2023-12-12-multi-modal-dataset-16]
- Multi-Modal Link 2 [www.kaggle.com/datasets/ramyharik/ff-2023-12-12-multi-modal-dataset-26]
- Multi-Modal Link 3 [www.kaggle.com/datasets/ramyharik/ff-2023-12-12-multi-modal-dataset-36]
- Multi-Modal Link 4 [www.kaggle.com/datasets/ramyharik/ff-2023-12-12-multi-modal-dataset-46]
- Multi-Modal Link 5 [www.kaggle.com/datasets/ramyharik/ff-2023-12-12-multi-modal-dataset-56]
- Multi-Modal Link 6 [www.kaggle.com/datasets/ramyharik/ff-2023-12-12-multi-modal-dataset-66]

# Acknowledgements

This work is funded in part by NSF Award 2119654 "RII Track 2 FEC: Enabling Factory to Factory (F2F) Networking for Future Manufacturing," and "Enabling Factory to Factory (F2F) Networking for Future Manufacturing across South Carolina," funded by South Carolina Research Authority. Any opinions, findings, conclusions, or recommendations expressed in this material are those of the author(s) and do not necessarily reflect the views of the sponsors.

# License

# Appendix

| Asset | Sensor Values | Data Type | Multi-Modal Dataset | Analog Dataset | Description |
|---|---|---|---|---|---|
| Conveyors | Q_VFD1_Temperature | Float | ✓ | ✓ | The temperature of conveyor 1 in Fahrenheit |
| | Q_VFD2_Temperature | Float | ✓ | ✓ | The temperature of conveyor 2 in Fahrenheit |
| | Q_VFD3_Temperature | Float | ✓ | ✓ | The temperature of conveyor 3 in Fahrenheit |
| | Q_VFD4_Temperature | Float | ✓ | ✓ | The temperature of conveyor 4 in Fahrenheit |
| | M_Conv1_Speed_mmps | Integer | ✓ | | The speed of the conveyor 1 in mm/s |
| | M_Conv2_Speed_mmps | Integer | ✓ | | The speed of the conveyor 2 in mm/s |
| | M_Conv3_Speed_mmps | Integer | ✓ | | The speed of the conveyor 3 in mm/s |
| | M_Conv4_Speed_mmps | Integer | ✓ | | The speed of the conveyor 4 in mm/s |
| Grippers | I_R01_Gripper_Pot | Integer | ✓ | ✓ | The analog output signal of the potentiometer on the Robot 1 gripper |
| | I_R02_Gripper_Pot | Integer | ✓ | ✓ | The analog output signal of the potentiometer on the Robot 2 gripper |
| | I_R03_Gripper_Pot | Integer | ✓ | ✓ | The analog output signal of the potentiometer on the Robot 3 gripper |
| | I_R04_Gripper_Pot | Integer | ✓ | ✓ | The analog output signal of the potentiometer on the Robot 4 gripper |
| | I_R01_Gripper_Load | Integer | ✓ | ✓ | The analog output signal of the load cell on the Robot 1 gripper |
| | I_R02_Gripper_Load | Integer | ✓ | ✓ | The analog output signal of the load cell on the Robot 2 gripper |
| | I_R03_Gripper_Load | Integer | ✓ | ✓ | The analog output signal of the load cell on the Robot 3 gripper |
| | I_R04_Gripper_Load | Integer | ✓ | ✓ | The analog output signal of the load cell on the Robot 4 gripper |
| Robot 1 | M_R01_SJointAngle_Degree | Float | ✓ | ✓ | The joint S angle of Robot 1 in degrees |
| | M_R01_LJointAngle_Degree | Float | ✓ | ✓ | The joint L angle of Robot 1 in degrees |
| | M_R01_UJointAngle_Degree | Float | ✓ | ✓ | The joint U angle of Robot 1 in degrees |
| | M_R01_RJointAngle_Degree | Float | ✓ | ✓ | The joint R angle of Robot 1 in degrees |
| | M_R01_BJointAngle_Degree | Float | ✓ | ✓ | The joint B angle of Robot 1 in degrees |
| | M_R01_TJointAngle_Degree | Float | ✓ | ✓ | The joint T angle of Robot 1 in degrees |
| Robot 2 | M_R02_SJointAngle_Degree | Float | ✓ | ✓ | The joint S angle of Robot 2 in degrees |
| | M_R02_LJointAngle_Degree | Float | ✓ | ✓ | The joint L angle of Robot 2 in degrees |
| | M_R02_UJointAngle_Degree | Float | ✓ | ✓ | The joint U angle of Robot 2 in degrees |
| | M_R02_RJointAngle_Degree | Float | ✓ | ✓ | The joint R angle of Robot 2 in degrees |
| | M_R02_BJointAngle_Degree | Float | ✓ | ✓ | The joint B angle of Robot 2 in degrees |
| | M_R02_TJointAngle_Degree | Float | ✓ | ✓ | The joint T angle of Robot 2 in degrees |
| Robot 3 | M_R03_SJointAngle_Degree | Float | ✓ | ✓ | The joint S angle of Robot 3 in degrees |
| | M_R03_LJointAngle_Degree | Float | ✓ | ✓ | The joint L angle of Robot 3 in degrees |
| | M_R03_UJointAngle_Degree | Float | ✓ | ✓ | The joint U angle of Robot 3 in degrees |
| | M_R03_RJointAngle_Degree | Float | ✓ | ✓ | The joint R angle of Robot 3 in degrees |
| | M_R03_BJointAngle_Degree | Float | ✓ | ✓ | The joint B angle of Robot 3 in degrees |
| | M_R03_TJointAngle_Degree | Float | ✓ | ✓ | The joint T angle of Robot 3 in degrees |
| Robot 4 | M_R04_SJointAngle_Degree | Float | ✓ | ✓ | The joint S angle of Robot 4 in degrees |
| | M_R04_LJointAngle_Degree | Float | ✓ | ✓ | The joint L angle of Robot 4 in degrees |
| | M_R04_UJointAngle_Degree | Float | ✓ | ✓ | The joint U angle of Robot 4 in degrees |
| | M_R04_RJointAngle_Degree | Float | ✓ | ✓ | The joint R angle of Robot 4 in degrees |
| | M_R04_BJointAngle_Degree | Float | ✓ | ✓ | The joint B angle of Robot 4 in degrees |
| | M_R04_TJointAngle_Degree | Float | ✓ | ✓ | The joint T angle of Robot 4 in degrees |
| Safety | I_SafetyDoor1_Status | Bool | ✓ | ✓ | "True" if Safety Door 1 is open and "False" if otherwise |
| | I_SafetyDoor2_Status | Bool | ✓ | ✓ | "True" if Safety Door 2 is open and "False" if otherwise |
| | I_HMI_EStop_Status | Bool | | ✓ | "True" if the HMI E-Stop button has been pressed and "False" if otherwise |
| Cycle Management | Q_Cell_CycleCount | Integer | ✓ | ✓ | Integer value representing the number of cycles elapsed. **<u>Note:</u>** This number resets to zero whenever the cycle was interrupted |
| Material Handling Station | I_MHS_GreenRocketTray | Bool | ✓ | ✓ | "True" if the Green Rocket Tray is detected in the Material Handling Station and "False" if otherwise |
| Stopper | I_Stopper1_Status | Bool | | ✓ | "True" if Stopper 1 is extended and "False" if otherwise |
| | I_Stopper2_Status | Bool | | ✓ | "True" if Stopper 2 is extended and "False" if otherwise |
| | I_Stopper3_Status | Bool | | ✓ | "True" if Stopper 3 is extended and "False" if otherwise |
| | I_Stopper4_Status | Bool | | ✓ | "True" if Stopper 4 is extended and "False" if otherwise |
| | I_Stopper5_Status | Bool | | ✓ | "True" if Stopper 5 is extended and "False" if otherwise |
| Cameras | Path1 | String | ✓ | | Path to image taken from Camera 1 |
| | Path2 | String | ✓ | | Path to image taken from Camera 2 |